\def\year{2020}\relax

\documentclass[letterpaper]{article} 
\usepackage{aaai20}  
\usepackage{times}  
\usepackage{helvet} 
\usepackage{courier}  
\usepackage[hyphens]{url}  
\usepackage{graphicx} 
\usepackage{graphicx}  
\usepackage[utf8]{inputenc}

\newcommand{\citet}[1]{\citeauthor{#1}
\shortcite{#1}}
\newcommand{\citep}{\cite}

\title{A Transfer Learning Method for Goal Recognition\\
Exploiting Cross-Domain Spatial Features}

\author{
 Thibault Duhamel, Mariane Maynard and Froduald Kabanza\\
 PLANIART \\
 Université de Sherbrooke \\
 \texttt{\{thibault.duhamel, mariane.maynard, froduald.kabanza\}@usherbrooke.ca} \\
}

\newtheorem{definition}{Definition}

\begin{document}
\maketitle

\begin{abstract}

The ability to infer the intentions of others, predict their goals, and deduce their plans are critical features for intelligent agents. For a long time, several approaches investigated the use of symbolic representations and inferences with limited success, principally because it is difficult to capture the cognitive knowledge behind human decisions explicitly. The trend, nowadays, is increasingly focusing on learning to infer intentions directly from data, using deep learning in particular. We are now observing interesting applications of intent classification in natural language processing, visual activity recognition, and emerging approaches in other domains. This paper discusses a novel approach combining few-shot and transfer learning with cross-domain features, to learn to infer the intent of an agent navigating in physical environments, executing arbitrary long sequences of actions to achieve their goals. Experiments in synthetic environments demonstrate improved performance in terms of learning from few samples and generalizing to unseen configurations, compared to a deep-learning baseline approach.

\end{abstract}

\section{Introduction}

Goal recognition is a critical feature of intelligent agents, as it allows them to anticipate future behaviors that have not yet been observed and integrate \textit{a priori} knowledge to make informed decisions. It is a fundamental cognitive ability lying at the heart of social interactions, often unconsciously, unlocking the possibility to understand beyond explicit communication. While humans intuitively manage to implicitly recognize and predict a course of action of others by observing them, granting machines such a strong ability remains a challenge.

This problem conveys several dimensions of complexity. Short-term action recognition, for instance, focuses on identifying activities over a short horizon using low-level sensors \cite{kong_fu_18,wang_etal_17}, as opposed to long-term goal recognition that aims to predict sequences of actions over a longer horizon~\cite{sukthankar_etal_14}. Behaviors can be fully or partially observable~\cite{keren_etal_16} and may involve multiple agents, cooperating or competing with each other~\cite{freedman_etal_17}.

In this paper, a single observer tries to infer the goal of a single agent evolving in a neutral environment, with a behavior either fully or partially observable. Traditional symbolic approaches to this type of goal recognition problem require handcrafted knowledge, engineered by experts, conveying the space of potential behaviors of the observed agent ~\cite{ramirez_geffner_10,masters_sardina_19a,pereira_etal_17,sohrabi_etal_16,vered_kaminka_17,vered_etal_18}. Unfortunately, it has proven difficult to express such knowledge in practice since a part of the human decision making is unconscious, hence impossible to model explicitly.

Using deep learning to learn from data is an attractive alternative approach, not impeded by the limitations of handcrafted models. Deep learning is used, for instance, to classify intents of utterances in natural language processing~\cite{chen_etal_19,wen_etal_17} or activities in video analysis~\cite{liu_etal_18}. The sequentiality is a common feature of these two types of applications, as natural language consists of sequences of words, whereas activities in video analysis are short sequences of low-level actions. In contrast, there have been so far fewer research efforts to apply deep learning to infer goals behind long sequences of actions. The rare existing approaches use conventional deep learning architectures such as convolutional or long-short term memory networks, with the difficulty of being able to generalize across domains~\cite{min_etal_16,amado_etal_18}. For example, while it is possible to predict the goals of others walking around in one particular scenario, the learned model does not apply to a completely different one, let alone a new one with fewer samples.

Following a similar inquiry to these previous approaches, this paper aims to demonstrate how it is possible to combine transfer learning and few-shot learning to infer the goal of agents engaged in long navigation behaviors in a physical environment. Transfer learning consists in reusing a model optimized for a specific domain as a starting point and somehow adapt it for another one. While quite well understood for many applications, including NLP~\cite{jia_etal_18}, image analysis~\cite{dongmei_etal_17}, and others~\cite{tan_etal_18}, this technique has, to the best of our knowledge, never been used for inferring the goal of agents engaged in long-term behaviors such as in the navigation domain. Few-shot learning, on the other hand, consists of optimizing a learner with a significantly reduced amount of examples~\cite{ravi_etal_17}. As far as we know, no paper applied this technique for long-term goal recognition.

Humans are efficient at learning to perform a variety of tasks in different domains, using a shared base of knowledge that is transferable (with perhaps a few examples). For instance, in video games, when a player faces a door, he intuitively searches for a key that might open it, leveraging related information gathered in his everyday experiences.

Likewise, we here use few-shot transfer learning to train a deep neural network in such a way that it would learn cross-domain intent patterns that are quickly adaptable to other scenarios in a navigation domain, with fewer training samples compared to baseline networks. To the best of our knowledge, no previous deep learning approach to goal recognition in the navigation domain (let alone any other domain conveying agents driven by long-term planning) has demonstrated an ability to generalize across different scenarios from a few examples.

To test this idea, we use a synthetic grid-world environment similar to previous approaches so far \cite{masters_sardina_19a}. However, instead of using a symbolic map as input for the goal recognizer, our approach utilizes raw bitmaps of higher resolution, similar to images or video-game deep learning applications~\cite{alphastarblog:19}. By doing so, the input framework in our approach is closer to real-world settings while being comparable to previous approaches.

We organize the rest of the paper as follows: first, we discuss the most relevant approaches while providing background concepts. Then, we describe our approach, followed by the experiments, along with a discussion.

\section{Related Work}

A large body of recent research for goal inference still uses symbolic knowledge representations and inferences. While we apply deep neural networks, it is useful to contrast both paradigms.

\subsection{Symbolic Goal Recognition}

Symbolic approaches to goal recognition traditionally cast inference processes as abductive reasoning, from observations to goals or plans, using some causal reasoning framework~\cite{sukthankar_etal_14}. A goal recognizer thus has two main components: (1) domain knowledge in some formal reasoning formalism, characterizing the potential behaviors of the observed agent; (2) an inference algorithm for reasoning about the knowledge, to infer goals or plans from observations.

The amount of knowledge required by these different approaches may vary. Most approaches request, in one way or another, knowledge about both the primitive actions of the observed agent and the rules governing its behaviors, also called plan libraries. Approaches based on Bayesian networks~\cite{charniak_goldman_93}, hidden Markov models~\cite{bui_etal_02}, Markov logic~\cite{song_etal_13}, hierarchical task networks (HTN)~\cite{avrahami_kaminka_05}, probabilistic grammars (which are in essence equivalent to HTNs augmented with a probabilistic model)~\cite{geib_goldman_09,kabanza_etal_13} can be grouped into that category. The so-called cost-based approaches or inverse-planning approaches only need a model of primitive actions of the observed agent, but not the plan library~\cite{ramirez_geffner_10,masters_sardina_19a,pereira_etal_17,sohrabi_etal_16,vered_kaminka_17,vered_etal_18}.

A common issue with all these approaches is the ability to represent the domain knowledge symbolically. Human beings, including experts, have difficulties accessing the unconscious mechanisms that participate in their decision making processes. It is a tremendous challenge for many domains -- those involving image recognition in particular -- to specify a knowledge-based model that can support goal inference in practice.

\subsection{Deep Learning of Models for Symbolic Goal Inference}

A natural thought about trying to overcome the knowledge-engineering challenge is to learn models for goal recognition. \citet{bisson_etal_15} experimented with the idea of using recursive neural networks to learn the probabilistic model underlying an HTN plan library used by a symbolic probabilistic goal recognizer. \citet{granada_etal_17} created, in a kitchen environment, a hybrid technique using a deep neural network to identify independent actions from sensors and the SBR algorithm (Symbolic Behavior Recognition) to recognize the goal achieved from the sequence of observations, with a plan library. \citet{pereira_etal_19} proposed to use a neural network to learn a nominal model (states and transition rules) of the environment and perform goal recognition with a planner on this model.

\subsection{End-to-End Deep-Learning for Goal Recognition}

This paper is interested in an end-to-end deep learning pipeline for goal recognition. Such an approach appears increasingly attractive, in the wake of recent breakthroughs solving complex games like Go~\cite{silver_etal_16} and real-time strategy games~\cite{alphastarblog:19}.

Various applications of video analysis show that deep learning is making significant inroads in recognizing short activities performed by people~\cite{liu_etal_18,wang_etal_17,kong_fu_18}. Nevertheless, the discussion here focuses on applications for long-term behaviors.

Long Short Term Memory networks (LSTM) have been used to recognize the goal of quite long-term behaviors by a player in the \textsc{CRYSTAL ISLAND} open-world game~\citet{min_etal_16,min_etal_17}. Using data collected from in-game interactions, an LSTM was trained to predict the player's goal from his sequence of interactions, with reliable performance.

\citet{amado_etal_18} introduced a pipeline to recognize the goal achieved by a player in different simple games (such as 8-puzzle and tower of Hanoi) from constructed images of the game state, divided into three steps. First, they convert inputs into a latent space (which is a representation of state features) using a dense auto-encoder network previously introduced in \citet{asai_fukunaga_18}. Then, an LSTM network utilizes this representation to perform a regression task, consisting of constructing a goal prediction in the latent space. Finally, a decoder network reconstructs the image of the goal from its latent representation.

While these learning architectures demonstrate impressive capabilities to perform goal recognition, they are unable to generalize to previously unseen configurations, as we shall illustrate with examples in the navigation domain. It is an essential point because an environment might change or slightly evolve (game updates, for instance). Another limitation, closely related, lies in the amount of data required to train a neural network.

\section{Proposed Method}

The fundamental motivation behind transfer learning is to reduce the effort needed to label new data for scenarios never seen beforehand, which is crucial in real-life applications. We believe similar domains must share identical features, as it is the case in image classification \cite{kopuklu_etal_19}: no matter what classes are to be recognized, there are always patterns like edges, lines, and other shapes involved in the learning process. From this point forward, it seems counterproductive to learn those features again for different targets.

Our approach for a deep learning framework with improved generalization capability and reduced training examples consists of combining transfer learning and few-shot learning. With a data representation tweak allowing to convey temporal information into a trajectory trail, we show that navigation goals can be inferred by a CNN alone, without an LSTM, along with a generalization ability to different navigation maps using reduced training samples.

\subsection{Deep Learning Architecture}

A convolutional neural network (CNN) is a specific deep learning architecture designed to exploit spatial proximity using the convolution operation. Filters of parameters are shared by translating a kernel across the dimensions of the input, especially advantageous with images or, in our case, 2D grids:

$$h_{x',y'} = \sigma(W*i) = \sigma(\sum_{x,y=0}^{M,N}{W_{x,y}}i_{x'-x,y'-y})$$

where $h$ is the output matrix, $\sigma$ is the activation function, $W$ is the kernel of size $(N,M)$, $*$ is the convolution operation and $i$ is an input window.

Our stacked data representation motivates our choice to use a CNN for goal recognition since the property of local connectedness is a convenient component to identify links between adjacent pixels, whether they are consecutive observations, walls, starts, or goals.

\subsection{Data Representation}

Projecting data to a subs-space (or latent space) to reduce the dimension by abstracting over irrelevant information that impairs the learning process is a technique often used in deep learning~\cite{pan_etal_08,zhao_etal_16}. 

In our approach, we represent a sequence of observations as a stacked spatial trail on a bitmap, thus projecting the temporal dimension on a 2D space. As far as we know, \citet{liu_etal_18,yan_etal_18} initially introduced this representation in short-term activity recognition, converting the time axis to a third spatial dimension for a 3D convolutional network, such that to exploit the closeness between two consecutive states for a given pixel.

In our navigation domain, we suspect there is a link between two successive events and two points of short distance, specifically in the navigation domain. In other words, the spatial closeness is equivalent to temporal proximity in the studied context, or at least sufficient to lose no equivalent information. Thus, in a stacked trajectory trail, the temporal information is conveyed by the trail, suggesting that a CNN, instead of a CNN combined with an LSTM, would be enough to learn the features relevant for goal recognition, saving us computation and memory resources. 

Given a sequence of observed positions $O=\{o_1=(x_1,y_1),..., o_n=(x_n,y_n)\}$, a list of obstacle coordinates $C$, a start position $S$ and a list of 10 possible goals $G$, we build a 5-channels bitmap $(B_{i,j})_{i,j \in [1,N]}$ where:

\begin{itemize}
    \item $B_{i,j} = (1,0,0,0,0)\textit{ if }(i,j) \in C$
    \item $B_{i,j} = (0,1,0,0,0)\textit{ if }(i,j) \in O$
    \item $B_{i,j} = (0,0,1,0,0)\textit{ if }(i,j) = S$
    \item $B_{i,j} = (0,0,0,1,0)\textit{ if }(i,j) \in G$
    \item $B_{i,j} = (0,0,0,0,1)\textit{ otherwise (navigable tiles) }$
\end{itemize}

The intuition that an image of a trajectory trail fed to a CNN might be enough to learn to infer the goal destinations of an observed agent, without the additional use of an LSTM, appears logical from a human cognition standpoint. When people usually depict a trajectory on a map, they intuitively draw lines representing a flattened version of their temporal reasoning, implicitly considering the time steps. That way, the goal recognition problem of an agent navigating in a map is transformed into recognizing motion patterns, which, as we demonstrate later, a convolutional network can learn to extract.

\subsection{Few-Shot Transfer Learning}

The adaptation process commences with a basic gradient optimization of a convolutional network on a single map, with a fixed configuration of start, obstacles, and goals, with several examples provided for every possible goal. We refer to this data as the base training set and base testing set, of respective sizes 16000 and 12800. The network, hence trained during five epochs, is called the "base network" and will be the one to adapt in the subsequent steps.

Our critical hypothesis is that there could be patterns (what we name cross-domain features) identified by the first layers of the base network, that are not involved in the mere goal recognition process. On the contrary, similarly to image classification, they would only recognize key edges, lines, shapes, or points somehow derived from the raw bitmap input. These features, of course, would not be specific to just one scenario and should be preserved when adapting the network to a new one.

From this assumption, we aim to quickly adapt the base network with only a few shots for an unseen configuration of obstacles, start, and goals (named transfer training set and transfer testing set). The same configuration is used in the transfer training set and the transfer testing set but is different from the one used in the base sets. The transfer training set contains $n$ shots, where a shot consists of one example per goal and observability (25\%, 50\%, 75\%, and 100\%, see next section). There are thus $4n|G|$ different examples in the transfer training set. The transfer testing set also contains 12800 examples to evaluate transfer performances. To adapt the base network, we freeze a certain amount of the first layers, which means only the last ones will be affected by a new gradient optimization using the transfer training set for three epochs only.

In the experiment section, we evaluate the performance of the adapted network depending on three hyperparameters that we tune: the number of locked layers, the number of shots provided, and the transfer learning rate. We work with the accuracy metric, which is the ratio of correct predictions over the total number of predictions performed. A prediction is correct when its highest assigned probability score corresponds to the real goal. In case of ties, a random draw applies.

\section{Experiments}

We conducted the experiments on the navigation domain, one of the benchmarks currently used by state-of-the-art goal recognition algorithms~\cite{masters_sardina_19a}. The problem consists in predicting the destination of an agent moving on a map, given its trajectory so far (observations). We downloaded 30 StarCraft maps from the MovingAI website~\cite{sturtevant_etal_12}\footnote{MovingAI Lab: https://movingai.com/} and downscaled them to 512x512 pixels, in which the agent can move up, down, left or right to reach one goal amongst a set of 10 possible ones, randomly sampled. As mentioned above, we aim to increase the input resolution so that our examples become more realistic than toy ones.

\begin{figure}[ht]
    \centering
    \includegraphics[width=\linewidth]{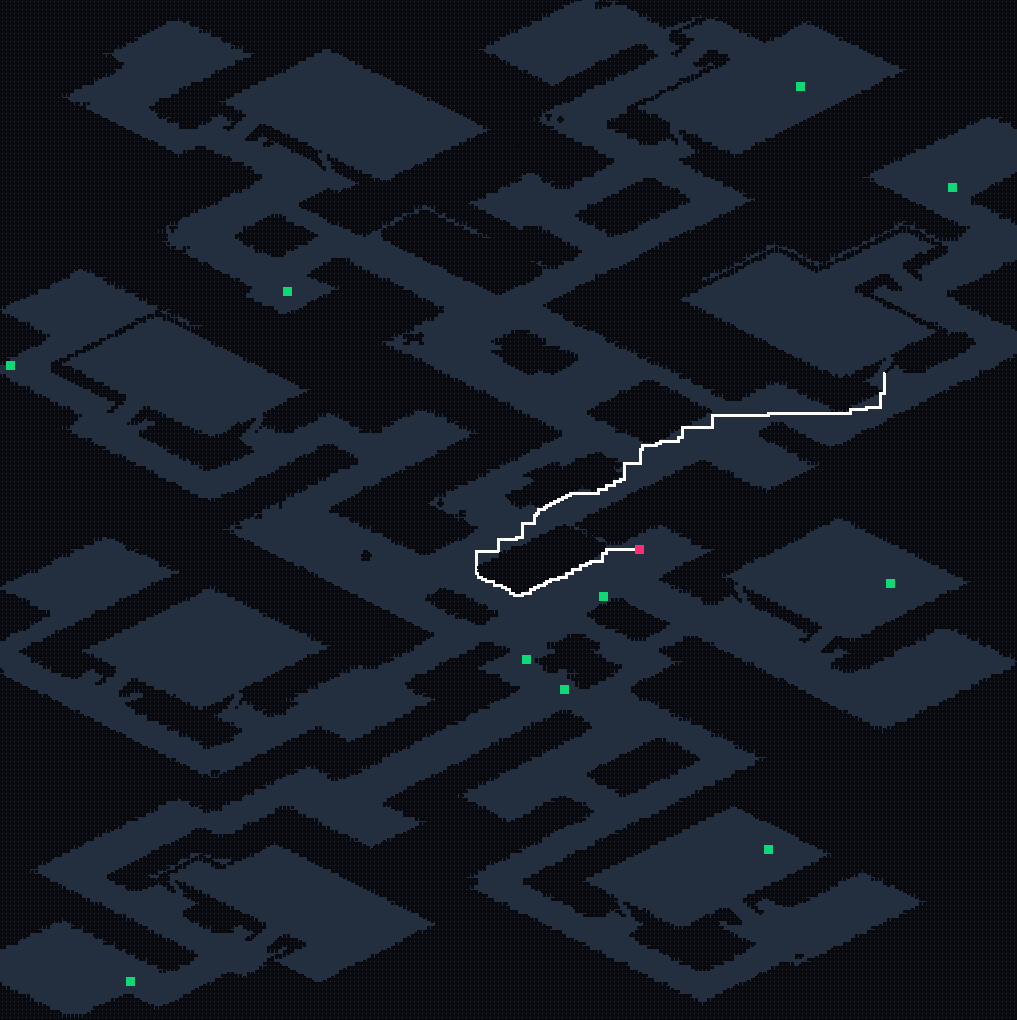}
    \caption{An input example fed to the network (512x512 pixels). 5 channels represent either a wall (black), a free tile (gray), an observation (white), a start (red) or a goal (green). The path is here truncated at 75\% of its total length. There are 10 possible goals.}
    \label{fig:75percent}
\end{figure}

Though still synthetic, we aim to tighten the frontier between generated and real-world data by using no handcrafted expert knowledge at all, providing as input for our networks just the pixels of the computed bitmaps (see figure \ref{fig:75percent}), made of 5 channels to represent either a navigable tile, an obstacle, an observation, a start or a goal. Moreover, we introduced noise in the agent's behavior by generating its path with a modified version of A* with a chance to drop an optimal step and pick a non-optimal one, to mimic a human-controlled route, using what we define as an $\epsilon$-over-estimating heuristic:
\begin{definition}
An \textit{$\epsilon$-over-estimating heuristic} is a function that returns an admissible quantity $h'$ with a chance of $1-\epsilon$, and $h' + \delta$ otherwise, where $\epsilon \in$ [0, 1] and $\delta > 0$.
\end{definition}
In practice, $\epsilon=0.2$ and $\delta=10$.

\begin{figure}[ht]
    \centering
    \includegraphics[width=\linewidth]{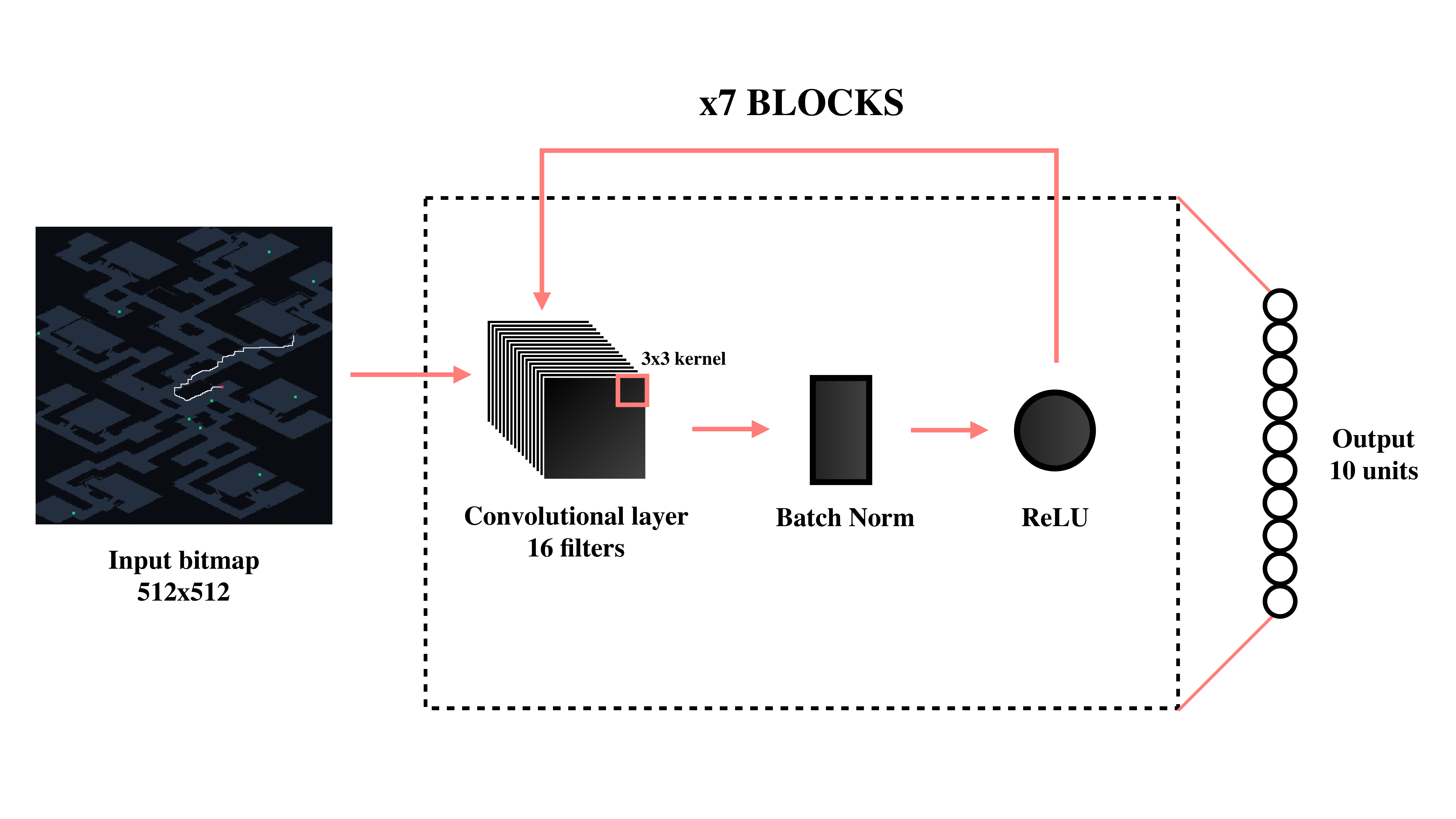}
    \caption{The architecture of our network.}
    \label{fig:network}
\end{figure}

We begin by verifying the key hypothesis that lies behind our transfer learning idea for goal recognition with a qualitative analysis. To validate the existence of a common base of knowledge, we initially trained a CNN network (figure \ref{fig:network}) on one configuration of the navigation domain and displayed its activation layers, from input to output. This CNN is a succession of 7 convolutional layers with 16 filters of 3x3 kernels interspersed with ReLU activations, followed by a final dense layer of 10 units with a softmax activation. It was optimized using Adam \cite{kingma_etal_14} with a learning rate of 0.01, $\beta_1=0.9$ and $\beta_2=0.999$, no decay, minimizing the cross-entropy loss.

\begin{figure}[ht]
    \centering
    \includegraphics[width=\linewidth]{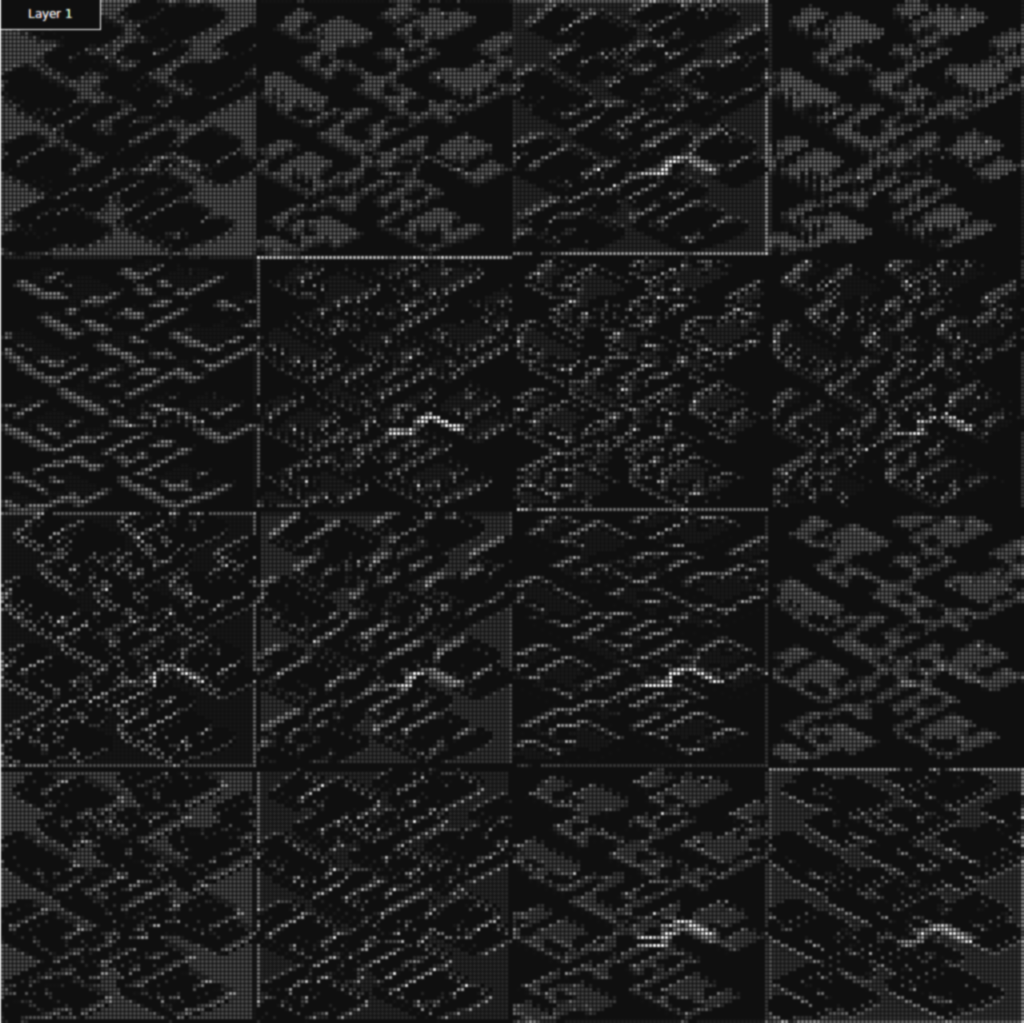}
    \caption[Visualisation des activations de la première couche cachée]{The activation of the first convolutional layer (16 filters). There are clear patterns of edges, free tiles, obstacles and observations.}
    \label{fig:layeractivation1}
\end{figure}

\begin{figure}[ht]
    \centering
    \includegraphics[width=\linewidth]{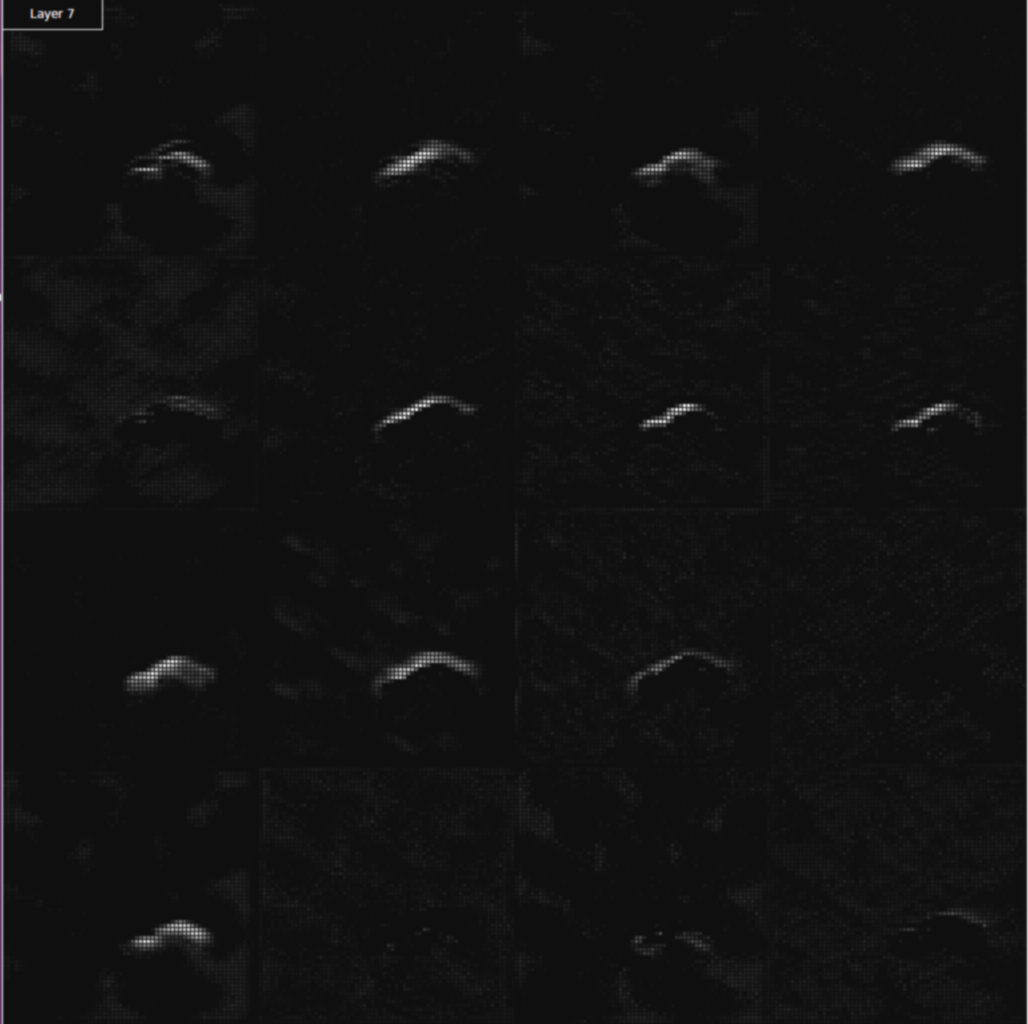}
    \caption{The activation of the last convolutional layer (16 filters). The network is now focusing on the trajectory trail and highlighting observations that seems to be important to recognize the goal.}
    \label{fig:layeractivation7}
\end{figure}

The images obtained from this visualization process (see figures \ref{fig:layeractivation1} and \ref{fig:layeractivation7}) suggest that the first layers handle feature extraction (such as observations, walkable areas, walls and edges between them), while the last ones perform the goal recognition task, at least visually. It is intriguing to observe that some filters do not even consider the trail of observations and only focus on the map configuration. In image classification, the same phenomenon reveals that the first layers often recognize edges, curves, and shapes. The pipeline thus progressively increases from low to high-level processing. From this observation, we plan to freeze layers that are not involved in the goal recognition process and re-train those who are.

We hence experimented with several transfer configurations and compared them with a network fully trained and tested on a single domain as a baseline. To begin with, we built a CNN slightly different from the one described above and trained it on a single map configuration. The purpose was to minimize dependencies between layers so that they would be easier to adapt. To do so, we replaced each convolutional layer by a block made of:

\begin{itemize}
    \item a convolutional layer (same parameters, with a He-Uniform initialization \cite{he_etal_15})
    \item a batch normalization for faster optimization and stability. It furthermore helps to reduce the impact of changing the input configuration.
    \item a ReLU activation
    \item a dropout chance of 0.1, to increase layer independence
\end{itemize}

We trained this base network during five epochs of 16000 examples, with paths truncated at 25\%, 50\%, 75\%, and 100\% of their full length (see figure \ref{fig:75percent}). In this paper, we do not consider full/partial observability as it is common in literature. Usually, an observability level of x\% denotes that x\% of the path is observed and taken into account. It means
there exists a chance that, for instance, an observability level of 1\% only retains the last positions, including the goal. Moreover, it is unadapted to online predictions because the full path is required beforehand. Here, we retain the first x\% of the path and study the convergence of our method, comparing online predictions at every step.

The classification target is simply a one-hot vector of size ten, and each output unit of the network is a probability score for each goal, measuring the certainty of its predictions.

We then duplicated this network, preserving its trained weights, and locked a certain number of layers. The remaining set of free weights was adapted to a new configuration, never seen before (different map, starting point, and goals locations). To do so, we introduce some transfer learning hyperparameters controlling the optimization process:

\begin{itemize}
    \item the number of shots required to adapt the network
    \item the number of layers that are frozen
    \item the transfer learning rate
\end{itemize}

The model was cross-validated by being adapted to 5 different new maps, with 3200 validation examples per map and per combination of hyperparameters. The mean accuracy for each hyperparameter value is studied below, according to the convergence metric just mentioned.

\subsection{Frozen Layers}

\begin{figure}[ht]
    \centering
    \includegraphics[width=\linewidth]{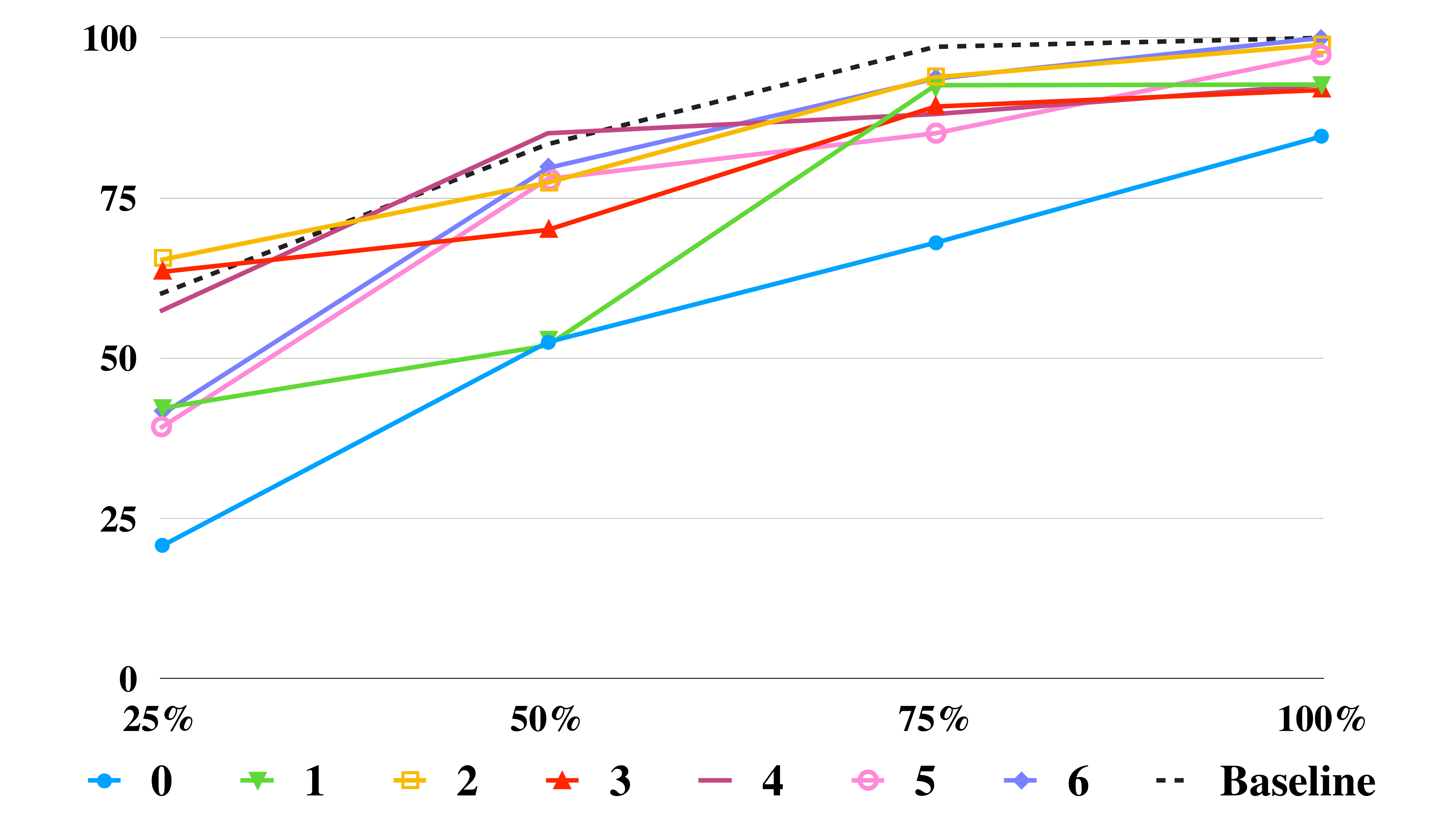}
    \caption{Average test accuracy of a network adapted to five unseen configurations, depending on the number of locked convolutional layers. The x-axis designates the percentage of observations retained from the complete path. The baseline network, trained and tested on a single configuration, is shown with the dashed line.}
    \label{fig:acc_vs_lock}
\end{figure}

The number of frozen layers may impact both the transfer learning quality and the adaptation duration. In this section, we set the number of shots to 5, along with a learning rate of 0.01. We selected those values after manually testing some configurations. Results are shown in figure \ref{fig:acc_vs_lock}.

There appears to be an optimal ratio to discover between the total number of layers and the number of layers to lock. When there are too many free layers (0 or 1 locked layer), the network performs poorly: since there are only a few shots available to train the free layers, a higher amount of weights should be far more challenging to optimize. Moreover, we assumed that the first layers were capable of handling features, and this experiment demonstrates that re-training those counteracts the adaptation of the entire set of weights.

The opposite is also inconvenient, as locking too many layers (5 or 6) hinders the tuning of the last goal recognition layers.

\subsection{Number of Shots}

The number of shots required to adapt a neural network is crucial in several contexts where critical classes are unbalanced. It is not the case here, but we may think of extreme cases (such as rare diseases and bomb attacks) where it is impossible to gather enough data to train a network from scratch. The human learning process is capable of fast convergence with just one example of a previously unseen entity, and it is a crucial feature we should grant to neural networks.

\begin{figure}[ht]
    \centering
    \includegraphics[width=\linewidth]{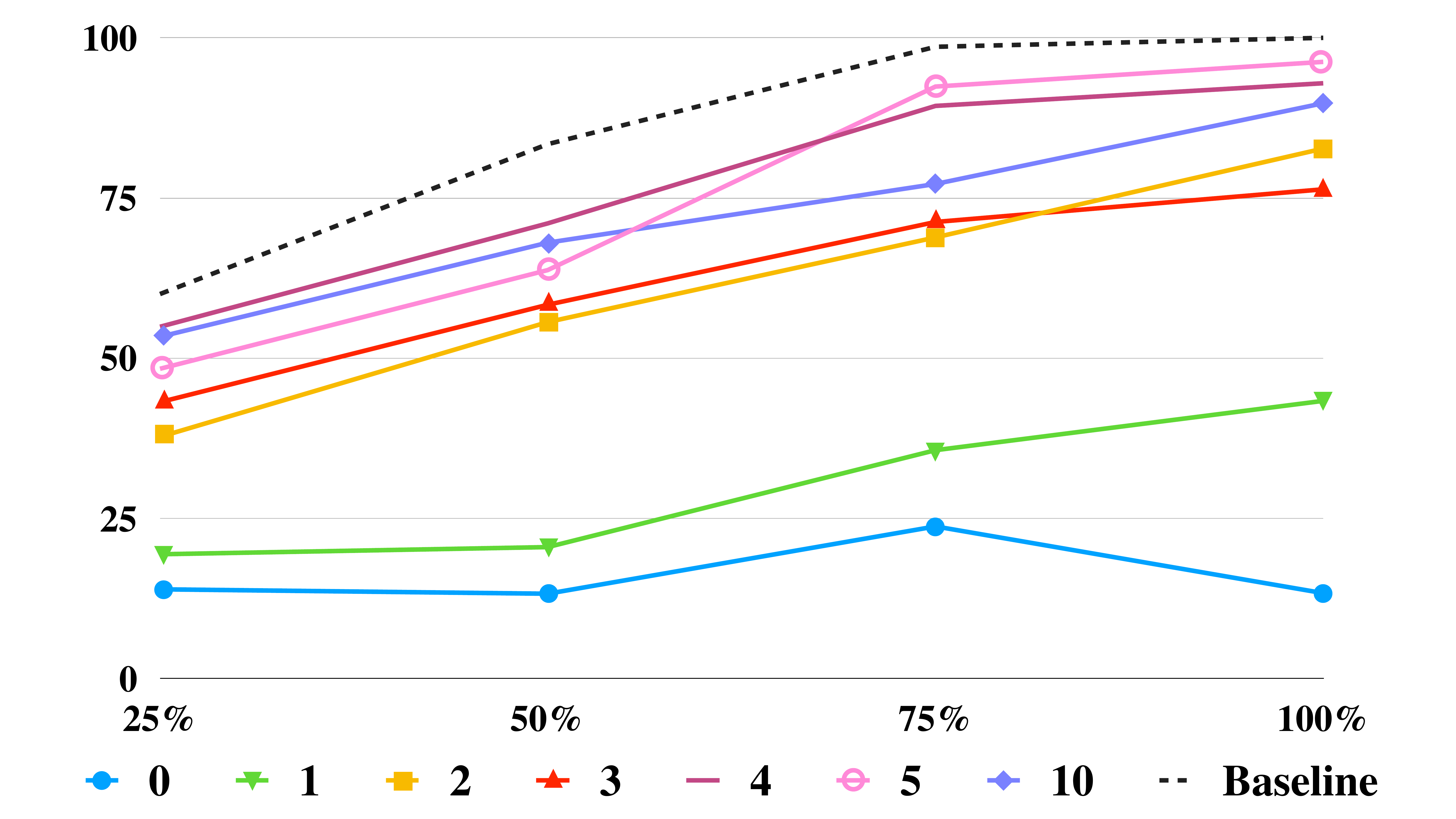}
    \caption{Average test accuracy of a network adapted to five unseen configurations, depending on the number of shots provided. The x-axis designates the percentage of observations retained from the complete path. The baseline network, trained and tested on a single configuration, is shown with the dashed line.}
    \label{fig:acc_vs_nbshots}
\end{figure}

In this section, we locked five layers in the base network (which means that two are free) and set the transfer learning rate to 0.01. The maximum number of shots we reached was ten since we noticed no improvement after this value. Figure \ref{fig:acc_vs_nbshots} summarizes the results for this experiment.

Results first illustrate that using the base network directly without adaptation (0 shot) does not show high performances and is not better than random noise, which means transfer learning is decisive.
The adapted network also poorly performs when provided with one shot, but already indicates excellent potential with 4 or 5 shots. It is almost as effective as if the complete network was fully trained from scratch and tested on the same map. However, we note that the network slightly overfits the small subset of examples when given too many shots (ten and above).

\subsection{Transfer Learning Rate}

The transfer learning rate controls how fast the weights of the network will converge when fed with new configurations. It is essential to tune this hyperparameter correctly, as there are only a few shots and epochs available. The key is to find a balance between underfitting, overfitting, converging, and diverging.

\begin{figure}[ht]
    \centering
    \includegraphics[width=1\linewidth]{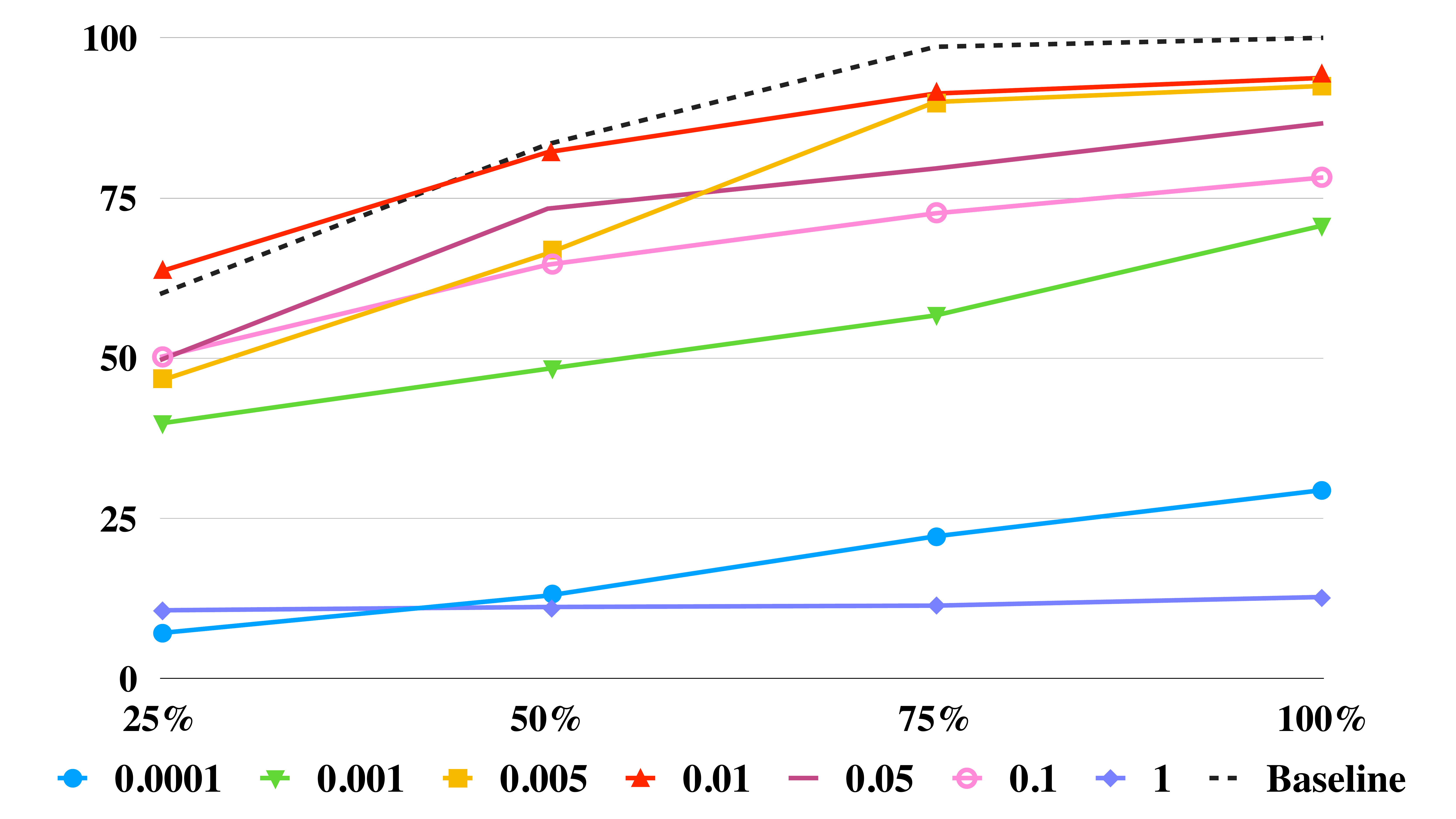}
    \caption{Average test accuracy of a network adapted to five unseen configurations, depending on the transfer learning rate. The x-axis designates the percentage of observations retained from the complete path. The baseline network, trained and tested on a single configuration, is shown with the dashed line.}
    \label{fig:acc_vs_lr}
\end{figure}

In this section, we set the number of shots to 5 and froze the first four layers. The results are shown in figure \ref{fig:acc_vs_lr}.

Low transfer learning rate values (0.001 and below) cause underfit issues, and high values lead to severe divergence (1 and above). The best value found is 0.01.

\bigskip

Global results reveal in the first place that our convolutional network built to recognize spatial patterns of behavior on projected bitmaps does perform effectively in the navigation domain. While this may not be the case in most environments where the temporal aspect is highly valuable, it is already interesting to notice how a different data representation, combined with an appropriate handling strategy, does not affect the prediction quality and can even compete with state-of-the-art literature results.

Moreover, our experimental benchmarks aimed to shrink the gap between synthetic and real data using only raw 512x512 pixels maps whose structure is akin to images. Hence, no bias is introduced by an expert exhibiting complex handcrafted rules and reconstructing a model of the environment, often impossible with real-world data.

However, applying our technique in real-life settings still requires a considerable amount of resources to collect sufficient data to train the base network in the first place.

\section{Conclusion}

We presented a few-shot transfer learning approach for goal recognition in the navigation domain, exploiting a new spatial trail representation with a convolutional network and providing only a few examples for fast weights adaptation. Our few-shot transfer learning method demonstrated great potential in a specific context of long-term behaviors, indeed assisting a standard deep-learning architecture in both generalizing and reasoning with visual features. We additionally experimented with high-resolution bitmaps as a step toward operating on real data.

We furthermore want to share the incentive that our method could be applicable in a variety of domains, real or synthetic, by identifying a shared, intermediate knowledge structure in the available data.

\section{Acknowledgements}

We thank \textit{Compute Canada} for the computing resources they provided to support our research project.

\bibliography{references}

\begin{thebibliography}{}

\bibitem[\protect\citeauthoryear{Amado \bgroup et al\mbox.\egroup
  }{2018}]{amado_etal_18}
Amado, L.; Aires, J.~P.; Pereira, R.~F.; Magnaguagno, M.~C.; Granada, R.; and
  Meneguzzi, F.
\newblock 2018.
\newblock Lstm-based goal recognition in latent space.
\newblock {\em CoRR} abs/1808.05249.

\bibitem[\protect\citeauthoryear{Asai and Fukunaga}{2018}]{asai_fukunaga_18}
Asai, M., and Fukunaga, A.
\newblock 2018.
\newblock Classical planning in deep latent space: Bridging the
  subsymbolic-symbolic boundary.
\newblock In {\em AAAI 2018}.

\bibitem[\protect\citeauthoryear{Avrahami{-}Zilberbrand and
  Kaminka}{2005}]{avrahami_kaminka_05}
Avrahami{-}Zilberbrand, D., and Kaminka, G.~A.
\newblock 2005.
\newblock Fast and complete symbolic plan recognition.
\newblock In {\em {IJCAI} 2005},  653--658.

\bibitem[\protect\citeauthoryear{Bisson, Larochelle, and
  Kabanza}{2015}]{bisson_etal_15}
Bisson, F.; Larochelle, H.; and Kabanza, F.
\newblock 2015.
\newblock Using a recursive neural network to learn an agent's decision model
  for plan recognition.
\newblock In {\em {IJCAI} 2015},  918--924.

\bibitem[\protect\citeauthoryear{Bui, Venkatesh, and West}{2002}]{bui_etal_02}
Bui, H.~H.; Venkatesh, S.; and West, G. A.~W.
\newblock 2002.
\newblock Policy recognition in the abstract hidden markov model.
\newblock {\em JAIR} 17:451--499.

\bibitem[\protect\citeauthoryear{Charniak and
  Goldman}{1993}]{charniak_goldman_93}
Charniak, E., and Goldman, R.~P.
\newblock 1993.
\newblock A bayesian model of plan recognition.
\newblock {\em Artif. Intell.} 64(1):53--79.

\bibitem[\protect\citeauthoryear{Chen \bgroup et al\mbox.\egroup
  }{2019}]{chen_etal_19}
Chen, C.; Zhang, X.; Ju, S.; Fu, C.; Tang, C.; Zhou, J.; and Li, X.
\newblock 2019.
\newblock Antprophet: an intention mining system behind alipay's intelligent
  customer service bot.
\newblock In {\em IJCAI 2019},  6497--6499.

\bibitem[\protect\citeauthoryear{Freedman and
  Zilberstein}{2017}]{freedman_etal_17}
Freedman, R.~G., and Zilberstein, S.
\newblock 2017.
\newblock Integration of planning with recognition for responsive interaction
  using classical planners.
\newblock In {\em AAAI}.

\bibitem[\protect\citeauthoryear{Geib and Goldman}{2009}]{geib_goldman_09}
Geib, C.~W., and Goldman, R.~P.
\newblock 2009.
\newblock A probabilistic plan recognition algorithm based on plan tree
  grammars.
\newblock {\em Artificial Intelligence} 173(11):1101--1132.

\bibitem[\protect\citeauthoryear{Granada \bgroup et al\mbox.\egroup
  }{2017}]{granada_etal_17}
Granada, R.; Pereira, R.; Monteiro, J.; Barros, R.; Ruiz, D.; and Meneguzzi, F.
\newblock 2017.
\newblock Hybrid activity and plan recognition for video streams.
\newblock In {\em {AAAI} 2017}.

\bibitem[\protect\citeauthoryear{Hana, Liu, and Fan}{2017}]{dongmei_etal_17}
Hana, D.; Liu, Q.; and Fan, W.
\newblock 2017.
\newblock A new image classification method using cnn transfer learning and web
  data augmentation.
\newblock {\em Expert Systems with Applications} 95.

\bibitem[\protect\citeauthoryear{He \bgroup et al\mbox.\egroup
  }{2015}]{he_etal_15}
He, K.; Zhang, X.; Ren, S.; and Sun, J.
\newblock 2015.
\newblock Delving deep into rectifiers: Surpassing human-level performance on
  imagenet classification.
\newblock {\em CoRR} abs/1502.01852.

\bibitem[\protect\citeauthoryear{Jia \bgroup et al\mbox.\egroup
  }{2018}]{jia_etal_18}
Jia, Y.; Zhang, Y.; Weiss, R.; Wang, Q.; Shen, J.; Ren, F.; Chen, z.; Nguyen,
  P.; Pang, R.; Lopez~Moreno, I.; and Wu, Y.
\newblock 2018.
\newblock Transfer learning from speaker verification to multispeaker
  text-to-speech synthesis.
\newblock In Bengio, S.; Wallach, H.; Larochelle, H.; Grauman, K.;
  Cesa-Bianchi, N.; and Garnett, R., eds., {\em Advances in Neural Information
  Processing Systems 31},  4480--4490.
\newblock Curran Associates, Inc.

\bibitem[\protect\citeauthoryear{Kabanza \bgroup et al\mbox.\egroup
  }{2013}]{kabanza_etal_13}
Kabanza, F.; Filion, J.; Benaskeur, A.~R.; and Irandoust, H.
\newblock 2013.
\newblock Controlling the hypothesis space in probabilistic plan recognition.
\newblock In {\em {IJCAI} 2013},  2306--2312.

\bibitem[\protect\citeauthoryear{Keren, Gal, and Karpas}{2016}]{keren_etal_16}
Keren, S.; Gal, A.; and Karpas, E.
\newblock 2016.
\newblock Goal recognition design with non-observable actions.
\newblock In {\em {AAAI} 2016}.

\bibitem[\protect\citeauthoryear{Kingma and Ba}{2014}]{kingma_etal_14}
Kingma, D.~P., and Ba, J.
\newblock 2014.
\newblock Adam: A method for stochastic optimization.
\newblock {\em CoRR} abs/1412.6980.

\bibitem[\protect\citeauthoryear{Kong and Fu}{2018}]{kong_fu_18}
Kong, Y., and Fu, Y.
\newblock 2018.
\newblock Human action recognition and prediction: {A} survey.
\newblock {\em CoRR} abs/1806.11230.

\bibitem[\protect\citeauthoryear{K{\"{o}}p{\"{u}}kl{\"{u}} \bgroup et
  al\mbox.\egroup }{2019}]{kopuklu_etal_19}
K{\"{o}}p{\"{u}}kl{\"{u}}, O.; Babaee, M.; H{\"{o}}rmann, S.; and Rigoll, G.
\newblock 2019.
\newblock Convolutional neural networks with layer reuse.
\newblock {\em CoRR} abs/1901.09615.

\bibitem[\protect\citeauthoryear{Liu \bgroup et al\mbox.\egroup
  }{2018}]{liu_etal_18}
Liu, K.; Liu, W.; Gan, C.; Tan, M.; and Ma, H.
\newblock 2018.
\newblock T-c3d: Temporal convolutional 3d network for real-time action
  recognition.
\newblock {\em AAAI}.

\bibitem[\protect\citeauthoryear{Masters and
  Sardi{\~{n}}a}{2019}]{masters_sardina_19a}
Masters, P., and Sardi{\~{n}}a, S.
\newblock 2019.
\newblock Cost-based goal recognition in navigational domains.
\newblock {\em JAIR} 64:197--242.

\bibitem[\protect\citeauthoryear{Min \bgroup et al\mbox.\egroup
  }{2016}]{min_etal_16}
Min, W.; Mott, B.~W.; Rowe, J.~P.; Liu, B.; and Lester, J.~C.
\newblock 2016.
\newblock Player goal recognition in open-world digital games with long
  short-term memory networks.
\newblock In {\em {IJCAI} 2016},  2590--2596.

\bibitem[\protect\citeauthoryear{Min \bgroup et al\mbox.\egroup
  }{2017}]{min_etal_17}
Min, W.; Mott, B.; Rowe, J.; Taylor, R.; Wiebe, E.; Boyer, K.; and Lester, J.
\newblock 2017.
\newblock Multimodal goal recognition in open-world digital games.
\newblock In {\em AAAI 2017}.

\bibitem[\protect\citeauthoryear{Pan, Kwok, and Yang}{2008}]{pan_etal_08}
Pan, S.~J.; Kwok, J.~T.; and Yang, Q.
\newblock 2008.
\newblock Transfer learning via dimensionality reduction.
\newblock In {\em Proceedings of the 23rd National Conference on Artificial
  Intelligence - Volume 2}, AAAI'08,  677--682.
\newblock AAAI Press.

\bibitem[\protect\citeauthoryear{Pereira \bgroup et al\mbox.\egroup
  }{2019}]{pereira_etal_19}
Pereira, R.~F.; Vered, M.; Meneguzzi, F.; and Ram{\'{\i}}rez, M.
\newblock 2019.
\newblock Online probabilistic goal recognition over nominal models.
\newblock In {\em Proceedings of the Twenty-Eighth International Joint
  Conference on Artificial Intelligence, {IJCAI} 2019, Macao, China, August
  10-16, 2019},  5547--5553.

\bibitem[\protect\citeauthoryear{Pereira, Oren, and
  Meneguzzi}{2017}]{pereira_etal_17}
Pereira, R.~F.; Oren, N.; and Meneguzzi, F.
\newblock 2017.
\newblock Landmark-based heuristics for goal recognition.
\newblock In {\em {AAAI} 2017},  3622--3628.

\bibitem[\protect\citeauthoryear{Ram\'{\i}rez and
  Geffner}{2010}]{ramirez_geffner_10}
Ram\'{\i}rez, M., and Geffner, H.
\newblock 2010.
\newblock Probabilistic plan recognition using off-the-shelf classical
  planners.
\newblock In {\em Proceedings of the Twenty-Fourth AAAI Conference on
  Artificial Intelligence}, AAAI'10,  1121--1126.
\newblock AAAI Press.

\bibitem[\protect\citeauthoryear{Ravi and Larochelle}{2017}]{ravi_etal_17}
Ravi, S., and Larochelle, H.
\newblock 2017.
\newblock Optimization as a model for few-shot learning.
\newblock In {\em 5th International Conference on Learning Representations,
  {ICLR} 2017, Toulon, France, April 24-26, 2017, Conference Track
  Proceedings}.

\bibitem[\protect\citeauthoryear{Silver \bgroup et al\mbox.\egroup
  }{2016}]{silver_etal_16}
Silver, D.; Huang, A.; Maddison, C.; Guez, A.; Sifre, L.; van~den Driessche,
  G.; Schrittwieser, J.; Antonoglou, I.; Panneershelvam, V.; Lanctot, M.;
  Dieleman, S.; Grewe, D.; Nham, J.; Kalchbrenner, N.; Sutskever, I.;
  Lillicrap, T.; Leach, M.; Kavukcuoglu, K.; Graepel, T.; and Hassabis, D.
\newblock 2016.
\newblock Mastering the game of go with deep neural networks and tree search.
\newblock {\em Nature} 529:484--489.

\bibitem[\protect\citeauthoryear{Sohrabi, Riabov, and
  Udrea}{2016}]{sohrabi_etal_16}
Sohrabi, S.; Riabov, A.~V.; and Udrea, O.
\newblock 2016.
\newblock Plan recognition as planning revisited.
\newblock In {\em IJCAI 2016}, IJCAI'16,  3258--3264.
\newblock AAAI Press.

\bibitem[\protect\citeauthoryear{Song \bgroup et al\mbox.\egroup
  }{2013}]{song_etal_13}
Song, Y.~C.; Kautz, H.~A.; Allen, J.~F.; Swift, M.~D.; Li, Y.; Luo, J.; and
  Zhang, C.
\newblock 2013.
\newblock A markov logic framework for recognizing complex events from
  multimodal data.
\newblock In {\em {ICMI} 2013},  141--148.

\bibitem[\protect\citeauthoryear{Sturtevant}{2012}]{sturtevant_etal_12}
Sturtevant, N.
\newblock 2012.
\newblock Benchmarks for grid-based pathfinding.
\newblock {\em Transactions on Computational Intelligence and AI in Games}
  4(2):144 -- 148.

\bibitem[\protect\citeauthoryear{Sukthankar \bgroup et al\mbox.\egroup
  }{2014}]{sukthankar_etal_14}
Sukthankar, G.; Geib, C.; Bui, H.~H.; Pynadath, D.; and Goldman, R.~P.
\newblock 2014.
\newblock {\em Plan, Activity, and Intent Recognition: Theory and Practice}.
\newblock San Francisco, CA, USA: Morgan Kaufmann Publishers Inc., 1st edition.

\bibitem[\protect\citeauthoryear{Tan \bgroup et al\mbox.\egroup
  }{2018}]{tan_etal_18}
Tan, C.; Sun, F.; Kong, T.; Zhang, W.; Yang, C.; and Liu, C.
\newblock 2018.
\newblock A survey on deep transfer learning.
\newblock In K{\r{u}}rkov{\'a}, V.; Manolopoulos, Y.; Hammer, B.; Iliadis, L.;
  and Maglogiannis, I., eds., {\em Artificial Neural Networks and Machine
  Learning -- ICANN 2018},  270--279.
\newblock Cham: Springer International Publishing.

\bibitem[\protect\citeauthoryear{Vered and Kaminka}{2017}]{vered_kaminka_17}
Vered, M., and Kaminka, G.~A.
\newblock 2017.
\newblock Heuristic online goal recognition in continuous domains.
\newblock In {\em {IJCAI} 2017},  4447--4454.

\bibitem[\protect\citeauthoryear{Vered \bgroup et al\mbox.\egroup
  }{2018}]{vered_etal_18}
Vered, M.; Pereira, R.~F.; Magnaguagno, M.~C.; Kaminka, G.~A.; and Meneguzzi,
  F.
\newblock 2018.
\newblock Towards online goal recognition combining goal mirroring and
  landmarks.
\newblock In {\em Proceedings of the 17th International Conference on
  Autonomous Agents and MultiAgent Systems, {AAMAS} 2018, Stockholm, Sweden,
  July 10-15, 2018},  2112--2114.

\bibitem[\protect\citeauthoryear{Vinyals \bgroup et al\mbox.\egroup
  }{2019}]{alphastarblog:19}
Vinyals, O.; Babuschkin, I.; Chung, J.; Mathieu, M.; and Jaderberg, M.
\newblock 2019.
\newblock {AlphaStar: Mastering the Real-Time Strategy Game StarCraft II}.

\bibitem[\protect\citeauthoryear{Wang \bgroup et al\mbox.\egroup
  }{2017}]{wang_etal_17}
Wang, J.; Chen, Y.; Hao, S.; Peng, X.; and Lisha, H.
\newblock 2017.
\newblock Deep learning for sensor-based activity recognition: A survey.
\newblock {\em Pattern Recognition Letters}.

\bibitem[\protect\citeauthoryear{Wen \bgroup et al\mbox.\egroup
  }{2017}]{wen_etal_17}
Wen, T.; Miao, Y.; Blunsom, P.; and Young, S.~J.
\newblock 2017.
\newblock Latent intention dialogue models.
\newblock In {\em Proceedings of the 34th International Conference on Machine
  Learning, {ICML} 2017, Sydney, NSW, Australia, 6-11 August 2017},
  3732--3741.

\bibitem[\protect\citeauthoryear{Yan, Xiong, and Lin}{2018}]{yan_etal_18}
Yan, S.; Xiong, Y.; and Lin, D.
\newblock 2018.
\newblock Spatial temporal graph convolutional networks for skeleton-based
  action recognition.
\newblock {\em AAAI}.

\bibitem[\protect\citeauthoryear{{Zhao} and {Du}}{2016}]{zhao_etal_16}
{Zhao}, W., and {Du}, S.
\newblock 2016.
\newblock Spectral–spatial feature extraction for hyperspectral image
  classification: A dimension reduction and deep learning approach.
\newblock {\em IEEE Transactions on Geoscience and Remote Sensing}
  54(8):4544--4554.

\end{thebibliography}
\bibliographystyle{aaai}
\end{document}